\pdfoutput=1

\documentclass[11pt,a4paper]{article}
\usepackage[final,hyperref]{naacl2021}
\usepackage{times}
\usepackage{amsmath}
\usepackage{bbm}
\usepackage{latexsym}
\usepackage{graphicx}
\usepackage{subcaption}
\usepackage{booktabs}
\usepackage{color,soul}

\usepackage{algorithm}
\usepackage[noend]{algpseudocode}

\usepackage{microtype}

\newenvironment{packed_enumerate}{
\begin{enumerate}
  \setlength{\itemsep}{1pt}
  \setlength{\parskip}{0pt}
  \setlength{\parsep}{0pt}
}{\end{enumerate}}

\newcommand{\scr}[1]{\mathcal{#1}}
\newcommand{\ind}{\mathbbm{1}}

\DeclareMathOperator*{\argmax}{arg\,max}

\title{Paragraph-level Simplification of Medical Texts}

\author{Ashwin Devaraj \\
  The University of Texas at Austin \\
  \texttt{ashwin.devaraj@utexas.edu} \\
  \\
  \textbf{Byron C.\ Wallace} \\
  Northeastern University \\
  \texttt{b.wallace@northeastern.edu} \\
  \And
  Iain J.\ Marshall \\
  King’s College London \\
  \texttt{iain.marshall@kcl.ac.uk} \\
  \\
  \textbf{Junyi Jessy Li} \\
  The University of Texas at Austin \\
  \texttt{jessy@austin.utexas.edu} \\
  }

\date{}

\begin{document}
\maketitle
\begin{abstract}
We consider the problem of learning to simplify medical texts.
This is important because most reliable, up-to-date information in biomedicine is dense with jargon and thus practically inaccessible to the lay audience. 
Furthermore, manual simplification does not scale to the rapidly growing body of biomedical literature, motivating the need for automated approaches.
Unfortunately, there are no large-scale resources available for this task.
In this work we introduce a new corpus of parallel texts in English comprising technical and lay summaries of all published evidence pertaining to different clinical topics.
We then propose a new metric based on likelihood scores from a masked language model pretrained on scientific texts. 
We show that this automated measure better differentiates between technical and lay summaries than existing heuristics.
We introduce and evaluate baseline encoder-decoder Transformer models for simplification and propose a novel augmentation to these in which we explicitly penalize the decoder for producing `jargon' terms; we find that this yields improvements over baselines in terms of readability.

\end{abstract}

\section{Introduction}

The need for accessible medical information has never been greater. 
A Pew Research survey of American's online health habits in 2013 revealed that ``one in three American adults have gone online to figure out a medical condition''~\cite{pew2013survey}.
Given the rise of medical misinformation on the internet~\cite{ioannidis2017survive},
accessibility has become an increasingly urgent issue~\cite{who2013health,armstrong2019counteracting}.
However, sources that provide accurate and up-to-date information, including scientific papers and \emph{systematic reviews} \cite{chalmers1995systematic}, are often effectively inaccessible to most readers because they are highly technical and laden with terminology~\cite{damay2006simtext}.

One potential solution to this problem is \emph{text simplification}, i.e., editing documents such that they are accessible to a wider audience, while preserving the key information that they contain.
Although manual simplification is too expensive to feasibly apply at scale, automatic text simplification~\cite{siddharthan2014survey,alva2020data} provides a potential means of rendering a large volume of specialist knowledge more accessible.

Large-scale data-driven simplification systems have mostly been trained on  Wikipedia~\cite{zhu2010monolingual,woodsend2011learning,coster2011simple} and news~\cite{xu2015problems}, and focus on sentence simplification~\cite{wubben2012sentence,wang2016text,xu2016optimizing,zhang2017sentence,kriz2019complexity,dong2019editnts,alva2020data}; on the other hand, medical text simplification is resource poor. Recent work has involved constructing sentence-aligned data automatically using monolingual text alignment methods~\cite{adduru2018towards,van2019evaluating}, but this process is noisy and constrains the task to sentence-level simplification.

\begin{table}
\centering
\small
\begin{tabular}{p{7.5cm}}
\toprule
\textbf{Technical abstract:}
Analysis showed a higher rate of weight gain in the high-volume feeds group: mean difference 6.20 g/kg/d (95\% confidence interval 2.71 to 9.69).
There was no increase in the risk of feed intolerance or necrotising enterocolitis with high-volume feeds, but 95\% confidence intervals around these estimates were wide. \\
\textbf{Plain-language summary:} 
Very low birth weight infants who receive more milk than standard volumes gain weight more quickly during their hospital stay.
We found no evidence suggesting that giving infants high volumes of milk causes feeding or gut problems, but this finding is not certain.\\
\bottomrule
\end{tabular}
\caption{Sample excerpts from a technical abstract (top) and corresponding plain-language summary (bottom) from the Cochrane Library.}
\label{tab:examples}
\end{table}

In this work we explore new data and modern conditional text generation models \cite{lewis2019bart} to simplify medical documents. We introduce a dataset of paired (technical, simplified) texts derived from the \href{https://www.cochranelibrary.com}{Cochrane Database of Systematic Reviews}, which is comprised of evidence syntheses on a wide range of clinical topics. Critically, each review includes a \emph{plain-language summary} (PLS) written by the authors. PLS are written directly from the full reviews with their own structure and guidelines; they are not simplified versions of the corresponding technical abstracts of the reviews, nor are they summaries of the abstracts.

However, we observe that portions of the PLS can be considered simplifications of analogous sections in the abstracts, that is, they contain roughly the same content but involve simplification operations such as paraphrasing, word/sentence deletion, and summarization. We heuristically derive 4459 such pairs of sections (or paragraphs) of technical--plain English bitexts.
We provide an excerpt of the dataset we have constructed in Table~\ref{tab:examples}.

This data allows us to explore characteristics of simplified versions of technical medical texts. We show that the differences in traditional readability metrics, such as Flesch-Kincaid~\cite{kincaid1975} and Automated Readability Index~\cite{ari1967}, are small. Instead, the differences are better captured using large-scale pre-trained masked language models, and this reveals that there is more to the language difference than the shallow cues such as sentence and word lengths that traditional readability metrics focus on.

We present baseline methods for automatic text simplification over this data and perform analyses that highlight the challenges of this important simplification task. We find that when naively fine-tuned for the task, existing encoder-decoder models such as BART~\cite{lewis2019bart} tend to prefer deletion over paraphrasing or explaining, and are prone to generating technical words. We propose a new approach to try and mitigate the latter issue by imposing a variant of unlikelihood loss~\cite{welleck2019neural} that explicitly penalizes the decoder for production of `technical' tokens. We show that this yields improvements in terms of readability with only a minor tradeoff with content quality.

In sum, this work takes a step towards paragraph-level simplification of medical texts by: (1) introducing a sizable new dataset, (2) proposing and validating a new masked language model (MLM)-based metric for scoring the technicality of texts, (3) analyzing and understanding the style of plain language in this important domain, and (4) presenting baselines that exploit a variant of unlikelihood training to explicitly penalize models for producing jargon. We release our code and data at \href{https://github.com/AshOlogn/Paragraph-level-Simplification-of-Medical-Texts} {\small \texttt{https://github.com/AshOlogn/Paragraph-\\level-Simplification-of-Medical-Texts}}.

\section{Related work}
Recent efforts on data-driven text simplification methods have tended to rely on two resources: the Wikipedia-Simple Wikipedia aligned corpus~\cite{zhu2010monolingual,woodsend2011learning,coster2011simple} and the Newsela simplification corpus~\cite{xu2015problems}. Yet, there is an urgent need to simplify medical texts due to health literacy levels \cite{who2013health}. However, due to a lack of resources with which to train model-based simplification systems in this domain, past work has tended to focus on lexical simplification~\cite{damay2006simtext,kandula2010semantic,abrahamsson-etal-2014-medical,mukherjee2017negait}.
Recently,~\citet{adduru2018towards} and~\citet{van2019evaluating} introduced sentence-aligned corpora at the scale of thousands of sentence pairs. In contrast to our corpus, these datasets were automatically derived using paraphrase mining or monolingual alignment processes. Furthermore, as these are exclusively sentence corpora, they limit the set of potential approaches to just those that operate over sentences. \citet{clear2019} created a simplification corpus for  medical texts in French, in which a small subset of the text pairs are manually sentence-aligned, resulting in 663 sentence pairs, 112 of which are also from Cochrane.

With respect to modeling, recent work has focused on sentence simplification, treating it as a  monolingual machine translation task~\cite{wubben2012sentence,wang2016text,xu2016optimizing} using encoder-decoder models~\cite{zhang2017sentence,kriz2019complexity,dong2019editnts}.  In the medical domain, existing systems tend to adopt lexical and syntactic simplification~\cite{damay2006simtext,kandula2010semantic,llanos2016}. Research on document simplification has been sparse; to the best of our knowledge, the few prior works on this in English have focused on analysis~\cite{petersen2007text}, sentence deletion~\cite{woodsend2011learning,zhong2020discourse}, and localized explanation generation~\cite{srikanth2020elaborative}. This work proposes and evaluates an encoder-decoder model for paragraph-level simplification.

\section{Technical abstracts vs.\ plain-language summaries}

We compiled a dataset of technical abstracts of biomedical systematic reviews and corresponding PLS from the Cochrane Database of Systematic Reviews, which comprises thousands of evidence synopses (where authors provide an overview of all published evidence relevant to a particular clinical question or topic). The PLS are written by review authors; Cochrane’s PLS standards~\cite{cochranePLS} recommend that ``the PLS should be written in plain English which can be understood by most readers without a university education''. PLS are not parallel with every sentence in the abstract; on the contrary, they are structured heterogeneously~\cite{kadic2016cochrane}.

\subsection{Data compilation}

To derive the dataset we scraped the online interface to the database for articles containing PLS, extracting the raw text of the technical abstracts and PLS for those that we identified. 
In this way we obtained 7820 pairs after removing problematic links (e.g., HTTP 404 errors).
We also excluded reviews with atypical formatting that would have required extensive manual inspection.

\begin{table}[]
\small
\centering
\begin{tabular}{llll}
    & & \multicolumn{2}{c}{\textbf{Compiled data}} \\
    \textbf{} & \textbf{Raw} & \textbf{Before-filter} & \textbf{After-filter} \\ \midrule
    Abstract  & $815\pm 331$  &   $551 \pm 272$                      & $501\pm 211$                 \\
    PLS       & $394\pm 216$ & $284\pm 156$                       & $264\pm 136$                
\end{tabular}
\caption{Means and standard deviations of original abstract and PLS lengths (tokens), and our compiled data before \& after filtering out texts with more than 1024 tokens.}
\label{tb:lendist}
\end{table}

On average, PLS are shorter than abstracts (Table \ref{tb:lendist}, `raw').
They contain sections different from those in the abstracts, emphasize different content, and sometimes contain information not in the abstract. We divided documents into those that are split into sections with subheadings and those without (henceforth ``long-form'' summaries); 56\% of the data are long-form. For the sectioned PLS, headers are quite different from those found in the abstracts. The latter adhere to one of the 2 following formats:
\begin{packed_enumerate}
\small
\item \emph{Background, Objectives, Search Methods, Selection Criteria, Data Collection and Analysis, Main Results, Authors’ Conclusions}
\item \emph{Background, Objectives, Methods, Main Results, Authors' Conclusions}
\end{packed_enumerate}
In contrast, PLS contain a variety of headings, with the most common ones shown below:
\begin{quote}
\small
\emph{background}, 
\emph{study characteristics}, 
\emph{key results},
\emph{review question},
\emph{quality of the evidence},
\emph{search date},
\emph{quality of evidence},
\emph{conclusions}
\end{quote}
Others include questions such as \emph{What was the aim of this review?} And \emph{How up-to-date was the review?}

Manual inspection revealed that the \emph{results}, \emph{discussion}, and \emph{conclusion} sections of abstracts and summaries tended to occur in parallel. This motivated us to extract aligned subsets of abstracts and summaries to compose our dataset. More specifically, we determined the approximate location of the section describing studies and results in each text and kept everything from that point forward.

Therefore, in the abstracts we kept the text from the \emph{Main Results} section onward. For the sectioned PLS we kept every section after and including the first that contained one of the following substrings: \emph{find}, \emph{found}, \emph{evidence}, \emph{tell us}, \emph{study characteristic}. For the long-form PLS, we found the first paragraph containing any of the following words within the first couple sentences and included that and subsequent paragraphs: \emph{journal, study, studies, trial}. We keep one-paragraph PLS in their entirety. We also exclude instances where the PLS and abstracts are drastically different in length, by keeping only instances where the length ratio between the two falls between 0.2 and 1.3.
Our final dataset comprises 4459 pairs of technical abstracts and PLS, all containing $\leq$1024 tokens (so that they can be fed into the BART model in their entirety).

\subsection{Characterizing readability differences}\label{sec:readability-metrics}

\paragraph{Readability metrics.}

Designing metrics that reliably capture readability remains an open topic of research. In recent years, a host of metrics have been developed that use a wide variety of linguistic features to assess readability in a supervised manner. For example, \citet{kate2010learning} developed a metric based on syntactical, semantic, and language model-based features, and \citet{vajjala2018} developed a new readability corpus, on which they trained support vector machines to predict text readability.
For this medical text simplification task, however, we considered a couple established heuristics-based readability metrics due to clear domain differences between our Cochrane corpus and those used to train supervised readability metrics: the Flesch-Kincaid score \cite{kincaid1975} and the automated readability index (ARI) \cite{ari1967}, which estimate the educational maturity (grade-level) required to comprehend a text. These metrics rely on a combination of shallow cues, most notably lengths of words, sentences, and documents.

Table~\ref{tb:readability_means} reports the mean grade levels of abstracts and PLS calculated via the above metrics. 
There are small but statistically significant ($p<0.01$, paired $t$-test) differences between the abstract and PLS distributions, especially for Flesch-Kincaid.
For instance, the maximum difference in mean minimum grades (1.5) is achieved by Flesch-Kincaid, and the number is only 0.6 with ARI.
By contrast, a 3--5 grade level difference was shown on the Wikipedia and Britannica simplification datasets~\cite{li2015specificity}.
The high grade-level suggested by standard readability metrics confirms prior studies highlighting that these `plain language' summaries of medical systematic reviews remain at higher reading levels than those of average US adults~\cite{karavcic2019languages}.

\begin{table}[]
\small
\centering
\begin{tabular}{lll}
    \textbf{Metric} & \textbf{Abstracts} & \textbf{PLS} \\ \midrule
    Flesch-Kincaid  & $14.4 \pm 2.3$                      & $12.9\pm 2.4$                 \\
    ARI             & $15.5\pm 2.8$                       & $14.9\pm 3.0$                
\end{tabular}
\caption{Means and standard deviations of different readability scores calculated over abstracts and PLS.}
\label{tb:readability_means}
\end{table}

\paragraph{Masked language models.}
Despite the small differences in readability metrics, PLS do qualitatively seem easier to understand (see Table~\ref{tab:examples} for an example). This suggests that existing measures are incomplete. 

We propose adopting modern masked language models --- namely BERT \cite{devlin2019bert} --- as another means of scoring the `technicality' of text. In particular, when such models are trained on specialized or technical language (e.g., scientific articles) we would expect the likelihoods subsequently assigned to `jargon' tokens to be relatively high compared to a model trained over general lay corpora, as in the original BERT model \cite{devlin2019bert}.

Capitalizing on this intuition, we consider two large-scale pre-trained masked language models: (1) BERT~\cite{devlin2019bert} trained on BooksCorpus \cite{zhu2015bookscorpus} and English Wikipedia; and (2) SciBERT~\cite{beltagy2019scibert}, trained on a sample of 1.14 million technical papers from Semantic Scholar~\cite{ammar2018construction} (mostly biomedical and computer science articles). 
Inspired by the original training objective for these models, we compute a probability score for a document by splitting it into sentences, masking 10 subsets of 15\% of the tokens in each sentence (exempting \texttt{CLS} and \texttt{SEP}), computing the likelihoods of the original tokens in the distributions output by the model in each masked position, and averaging these probabilities over all the masked subsets and sentences in the document. The details are shown in Algorithm~\ref{alg:masked_prob}.

\begin{algorithm}
\caption{Used to compute a probability score for a text document $D$ given a masked language model $M$. The output of the model returned by a call to {\sc Forward} is a matrix where each row maps to a distribution over all the tokens in the vocabulary. The {\sc Append} function adds a value to the end of a list.}\label{alg:masked_prob}
\begin{algorithmic}
\Procedure{Masked-Prob}{$D, M$}
\State sents $\gets \Call{Sentence-Split}{D}$
\State $P\gets$  Initialize empty list
\For{$i=1\ldots |\textrm{sents}|$}
    \State $T\gets \Call{Tokenize}{\textrm{sents}[i]}$
    \For{$j=1\ldots 10$}
        \State $A \leftarrow$ sample 15\% from $1\ldots |T|$
        \State $T'\gets T$
        \ForAll{$a\in A$}
            \State $T'[a]\gets \textrm{{\tt [MASK]}}$
        \EndFor
        \State outputs $\gets \Call{Forward}{M, T'}$
        \ForAll{$a\in A$}
            \State prob $\gets {\textrm{outputs}}[a][T[a]]$
            \State $\Call{Append}{P,\: \textrm{prob}}$
        \EndFor
    \EndFor
\EndFor
\State 
\Return $\textrm{mean}(P)$
\EndProcedure
\end{algorithmic}
\end{algorithm}

\begin{figure}
  \centering
  \begin{minipage}[b]{0.25\textwidth}
    \includegraphics[width=\textwidth]{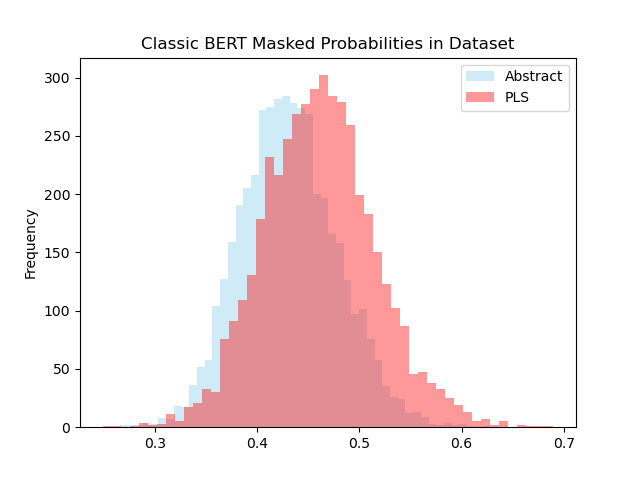}
  \end{minipage}%
  \begin{minipage}[b]{0.25\textwidth}
    \includegraphics[width=\textwidth]{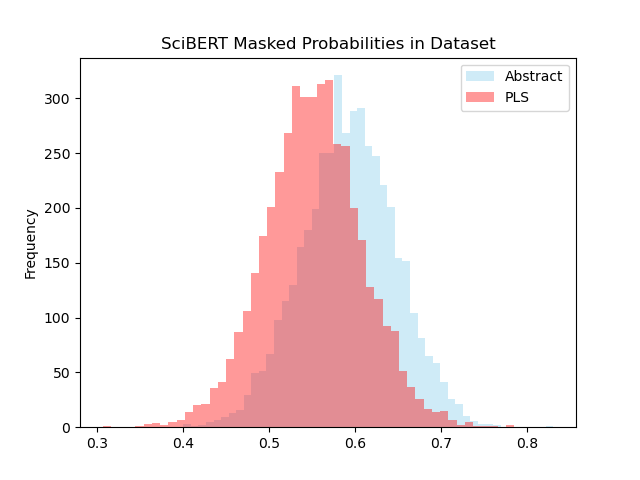}
  \end{minipage}
  \caption{BERT (left) vs SciBERT (right) probabilities of technical abstracts (blue) and PLS (red).}
  \label{fig:berts}
\end{figure}

Figure~\ref{fig:berts} depicts the distributions of probabilities output by general BERT and SciBERT for the abstracts and PLS in our dataset. 
Both masked LMs induce distributions over instances from the respective sets that are clearly different. 
For example, SciBERT (which yields sharper differences) outputs higher likelihoods for tokens comprising the technical abstracts than for those in the plain language versions, as we might expect given that this is pretrained on technical literature. 
A paired $t$-test confirms that these observed differences between the abstracts and PLS distributions are statistically significant (with $p < 0.01$). 

\paragraph{Which metric discriminates better?}

To better determine how well the proposed masked probability outputs discriminate between technical abstracts and PLS, we plot receiver operating characteristic (ROC) curves for the outputs of BERT, SciBERT, Flesch-Kincaid and ARI, coding technical and PLS abstracts as 0 and 1, respectively. The SciBERT curve has a higher AUC score (0.70) than the general BERT curve (0.66), indicating that it is better at discriminating between plain language and technical abstracts. For this reason, we use the SciBERT masked probabilities when analyzing the texts generated by our models.

The AUC score for SciBERT is also higher than that for Flesch-Kincaid, indicating that simplicity in PLS can be better captured by probabilistic means than by surface-level linguistic cues, and that it is more appropriately viewed as a stylistic difference rather than one of readability. This echoes the arguments made by early investigators of readability metrics that these measures do not replace more subtle linguistic characteristics, e.g., style~\cite{klare1963measurement,chall1958readability}.

\begin{figure}
  \centering
  \begin{minipage}[b]{0.5\textwidth}
    \includegraphics[width=\textwidth]{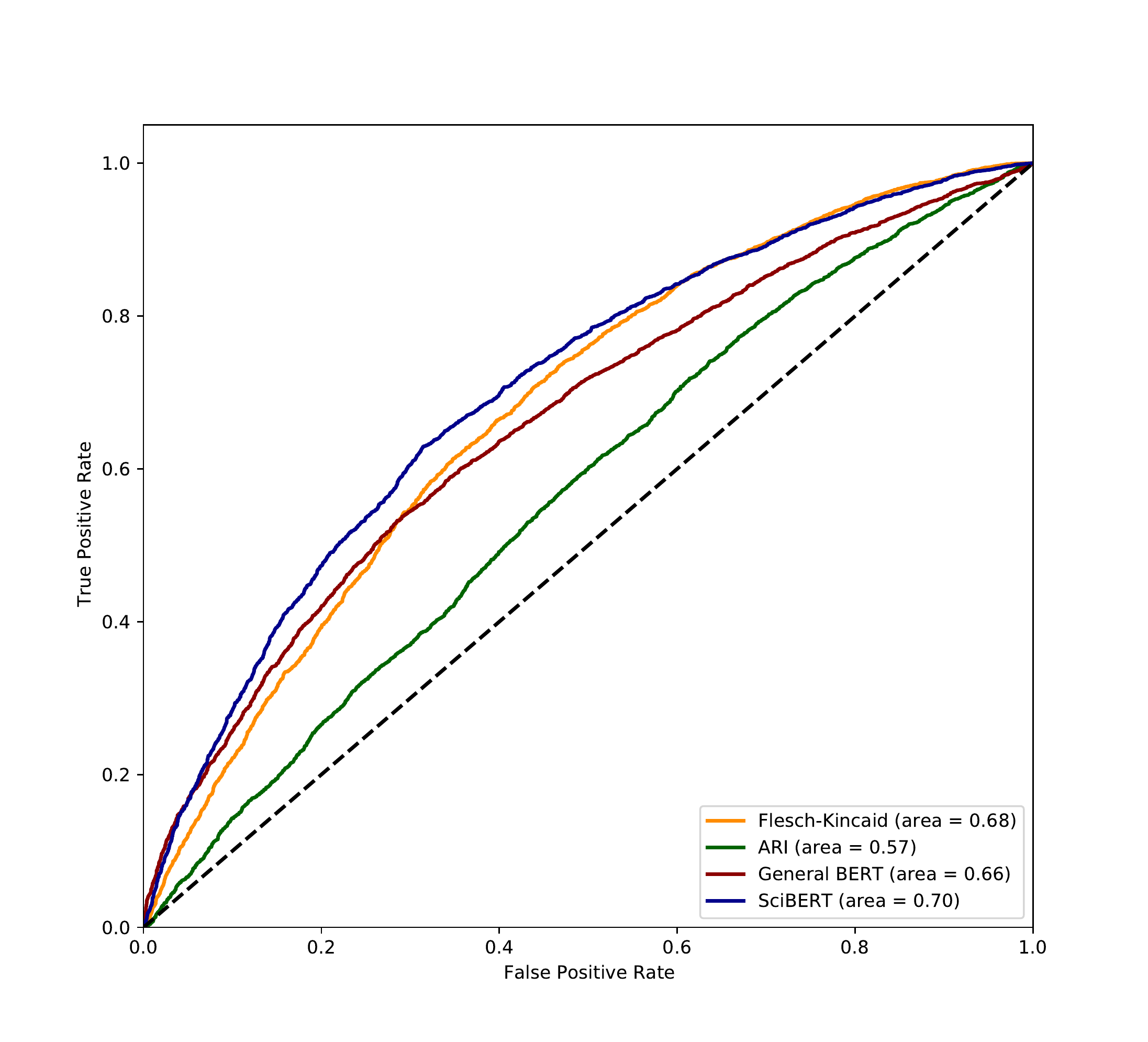}
  \end{minipage}
  \caption{ROC Curves for Readability Metrics.}
  \label{fig:readability_roc}
\end{figure}

\subsection{Lexical analysis}\label{sec:lexical-analysis}
We next investigate lexical differences between technical abstracts and PLS. 
In prior work, \citealt{gledhill2019} performed extensive lexical analysis on this corpus by comparing the relative frequencies of different part-of-speech $n$-grams found in the abstracts and PLS. 
Here, we analyze the weights from a logistic regression model that classifies whether a text is a technical abstract or a PLS (coding the latter as $y=1$); the weights learned by the model can be conveniently incorporated into the loss function we use to train our simplification model (Section \ref{sec:ull}).

We represent texts as normalized bag-of-words frequency vectors (with a feature for each token in the BART vocabulary). We performed 5-fold cross validation on the data and observed an average accuracy of 92.7\%, which indicated that even this relatively simple model is capable of accurately distinguishing technical abstracts from PLS. 
We also evaluated this model on the train-validation split described in Section \ref{sec:data}. The model achieves a very high AUC score of 0.99, indicating that it almost perfectly separates abstracts from PLS.

To better understand which kinds of tokens are most associated with technical abstracts and PLS, we examined the tokens with the highest-magnitude learned weights in the model, with the most negative weights corresponding to tokens indicative of technical abstracts and the most positive ones being indicative of PLS. These notable tokens are displayed in Table~\ref{tb:logr_weights}. From this table it is clear that numerical tokens and those related to statistical analysis, like \emph{bias} and \emph{CI} (confidence interval) are most indicative of abstracts. The tokens indicative of PLS are less illuminating and merely reflect common phrases include in PLS, such as \emph{In this review} and \emph{We searched scientific databases}.

In Section~\ref{sec:methods}, we use this model as a discriminator along with our transformer encoder-decoder model during training to penalize the generation of tokens that are indicative of technical abstracts.

\begin{table}[t]
\small
\centering
\begin{tabular}{ll|ll}
\textbf{Token} & \textbf{Weight} & \textbf{Token} & \textbf{Weight} \\ \midrule
0              & $-7.262$        & people         & $4.681$         \\
.              & $-6.126$        & review         & $4.551$         \\
\%             & $-5.379$        & We             & $4.461$         \\
CI             & $-4.986$        & This           & $3.413$         \\
;              & $-4.821$        & that           & $2.943$         \\
95             & $-4.593$        & The            & $2.836$         \\
significant    & $-4.273$        & side           & $2.722$         \\
R              & $-3.726$        & who            & $2.671$         \\
1              & $-3.685$        & blood          & $2.515$         \\
There          & $-3.477$        & found          & $2.514$         \\
bias           & $-3.303$        & searched       & $2.407$         \\
criteria       & $-3.263$        & The            & $2.114$         \\
outcome        & $-3.247$        & results        & $2.098$         \\
(              & $-3.195$        & their          & $2.022$         \\
inclusion      & $-3.148$        & current        & $1.984$        
\end{tabular}
\caption{The tokens with the most negative and most positive weights in a logistic regression model trained to distinguish technical abstracts from PLS.}
\label{tb:logr_weights}
\end{table}

\section{Baseline models for simplification}\label{sec:methods}

\subsection{Pretrained BART}

Our baseline simplification model is BART \citep{lewis2019bart}, an encoder-decoder architecture 
in which both components are transformers \cite{transformer}. The decoder is auto-regressive, making it a natural fit for generation tasks.
BART has been shown to achieve strong performance on text summarization, specifically on the CNN/Daily Mail \citep{hermann2015cnndm} and XSum \citep{narayan2018dont} datasets. 

We initialize the weights in BART to those estimated via fine-tuning on the XSum \cite{narayan2018dont} dataset as provided by HuggingFace's Model Hub \citep{Wolf2019HuggingFacesTS}. We then fine-tune these models on our corpus.\footnote{We also considered starting from a checkpoint corresponding to training over CNN/Daily News but preliminary manual examination of model outputs suggested starting from XSum yielded higher quality outputs.}

In the decoding step, we use nucleus sampling~\cite{holtzman2019curious}: at each step of token generation the next token is sampled from a probability distribution constructed by removing the `tail' of probability mass from BART's output distribution and then renormalizing. This strategy mitigates the awkward repetition typical of greedy methods like beam search while still avoiding incoherence by truncating the unlikely tail in the original model distribution.

\subsection{Unlikelihood training}\label{sec:ull}

As an additional mechanism to encourage simple terminology in the PLS generated by our model, we propose a new method in which we explicitly penalize the model for producing seemingly technical words via \emph{unlikelihood training} \citep{welleck2019neural,li2020dont}. The idea is to add a term to the objective that encourages the model to \emph{decrease} the probability mass assigned to some set of tokens $\mathcal{S}$. This is realized by adding a term to the (log) loss: $UL=\sum_{j=1}^{|\mathcal{S}|} -\log (1 - p_\theta(s_j|y_{<t}, x))$, where $x$ is the technical abstract input to the encoder, $y_{<t}$ is the prefix of the target summary $\mathbf{y}$ input to the decoder at time $t$, and $p_\theta(s_j|y_{<t}, x)$ is the probability assigned to token $s_j$ in the distribution output by BART (with model parameters $\theta$) at time $t$. This expression is referred to as \emph{Unlikelihood Loss} (UL). The UL term is weighted by a positive constant $\alpha$ and added to the typical log-likelihood objective.

We construct $\mathcal{S}$ by collecting tokens with negative weights from a bag-of-words logistic regression model trained to classify whether a document is simple (1) or complex (0), for which negative tokens are indicative of complex language. We then softmax the absolute values of these weights so that they sum to 1 and the tokens most indicative of technical abstracts (i.e., those with the most negative weights initially) contribute the most to this sum. 
We consider three variants of this procedure. (1) We classify whether a document is a PLS or an abstract (Section~\ref{sec:lexical-analysis}). 
(2) We use external data, namely the Newsela corpus~\cite{xu2015problems}, and train a model to distinguish between documents of reading levels 0 and 3.\footnote{Five-fold evaluation showed that the model achieved $>90$\% accuracy. We also experimented with the Simple Wikipedia/Wikipedia dataset~\cite{zhu2010monolingual}, but this model was not effective in early experiments.} (3) We train two different models for the previous tasks and then sum the weight vectors before applying a softmax to derive token penalties.

Let $w_j$ denote the learned logistic regression weight for token $s_j\in \mathcal{S}$. The final weight $w'_j$ used in the unlikelihood loss function is:

\begin{equation}
w'_j = \frac{\exp(|w_j|/T)}{\sum_{i=1}^{|\mathcal{S}|} \exp(|w_i|/T)}
\end{equation}

\noindent where $T$ is the temperature of the softmax.

A modification we make to the unlikelihood loss function is that we only apply the loss for a given token $s_j$ if the probability distribution output for the token at position $t$ indicates that $s_j$ should be output, that is, if $s_j = \underset{v\in \scr{V}}{\argmax}\:\: p_\theta(v | y_{< t})$ where $\scr{V}$ denotes BART's token vocabulary. Denoting an indicator function for this event by $\ind_{s_j, t}$, our final unlikelihood loss term $\scr{L}(p_\theta, \mathcal{S}, \mathbf{y})$ is: 

\begin{equation}
  -\sum_{t=1}^{|\mathbf{y}|} \sum_{j=1}^{|\mathcal{S}|} \ind_{s_j,t} w'_j \log(1-p_\theta(s_j|y_{< t}))
\end{equation}

\subsection{Experimental setup}\label{sec:data} \paragraph{Data.} We split our dataset of 4459 abstract-PLS pairs so that 3568 reviews are in the training set, 411 in the validation set, and 480 in the test set. 
We experimented with hyperparameters by manually inspecting a subset of the validation set and report results on the entire test set.

\paragraph{Hyperparameters.} For nucleus sampling, we use a top-$p$ value of 0.9.
In the unlikelihood training procedure, we experimented with different values of $\alpha$ in our total loss function ($1,10,10^3,10^6$) on the validation set and different temperatures $T$ in the softmax step ($1, 2, 5, 10$). Based on manual examination of the generated texts in the validation set, we determined that $(T=2, \alpha=100)$ yields the most coherent and high-quality simplifications, so we only report results for this case. All models are fine-tuned on our dataset for 1 epoch with a batch size of 1 and a learning rate that starts at 3e-5 and decreases linearly to 0 over the course of training. For optimizer, we used AdamW with $\varepsilon$ = 1e-8 ~\cite{adam, weightdecayreg}.

\section{Results}
In this section we comment on the generated texts' readability, quality of summarization and simplification, stylistic fidelity with the PLS, and overall coherence and simplicity based on human examination. 
In the results tables, we indicate whether lower or higher scores for the metrics reported are better with $\downarrow$ and $\uparrow$ symbols, respectively.

\subsection{Readability scores}

Table \ref{tab:readability_results} reports the mean readability scores achieved under different training settings. Results generated via models trained with the proposed UL objective achieve significantly lower Flesch-Kincaid scores than those achieved by both the technical abstracts and reference PLS, whereas the model trained without UL produced texts with a higher reading level than the PLS. Rather surprisingly, the UL-Newsela and UL-both settings, both of which use the Newsela dataset to produce unlikelihood weights, did not yield a decrease in estimated grade levels. We suspect that this could be attributed to the difference in domains, that is, the tokens contributed by the Newsela classifier are not generated frequently enough to have a noticeable impact during unlikelihood training.

These results suggest that: (1) BART is capable of performing simplification of medical texts such that outputs enjoy reduced reading levels compared to those of the technical abstracts; (2) The proposed use of UL to explicitly penalize the model for outputting jargon allows for the generation of text with even greater readability than the reference PLS. The reading levels of even the simplified outputs, however, are at the late-high school/early college levels. 
This could reflect the relatively small differences in readability scores between abstracts and PLS in general (Section~\ref{sec:readability-metrics}).

\subsection{Style}

In Section \ref{sec:readability-metrics} we showed that SciBERT masked probability scores are more useful as a discriminator between technical abstracts and PLS than the standard readability metrics, which use surface-level cues like word and sentence counts. Experiments by \citet{jawahar2019} suggest that BERT-style masked language models encode a wide array of syntactic and semantic features of language, which they then employs for downstream tasks. For this reason, we use SciBERT masked probability scores as our notion of style, with lower scores corresponding to simpler, less technical language.
To explore the extent to which the generated summaries stylistically resemble the PLS, we computed the average of the SciBERT masked probability scores of the generated texts for each model. The results are shown in Table \ref{tab:readability_results} along with the readability scores.

We see that every model produces text with significantly lower probability scores than the abstracts, which suggests that they successfully convert input abstracts into less-technical summaries. Though the average scores are higher than that of the PLS, this difference is not statistically significant, so we can consider the outputs of the models to be stylistically on par with the target PLS. 

\begin{table}[t]
\centering
\small
\begin{tabular}{lllll}
                 & \textbf{FK}$\downarrow$ & \textbf{ARI}$\downarrow$ & \textbf{SciBERT}$\downarrow$ \\
\midrule
\emph{Abstracts} & 14.42       & 15.58   & 0.57      \\
\emph{PLS}       & 13.11       & 15.08   & 0.53      \\
\midrule
No UL           & 13.44       & 15.09   & 0.55      \\
UL-Cochrane     & 11.97       & 13.73   & 0.55      \\
UL-Newsela      & 12.51       & 14.15   & 0.54      \\
UL-Both         & 12.26       & 14.04   & 0.54      \\
\end{tabular}
\caption{Flesch-Kincaid, ARI, and SciBERT masked probability scores for generated PLS. Differences wbetween abstracts and generated PLS are statistically significant; so are differences in FK and ARI between UL models and No-UL ($p<0.01$, paired $t$-test).}
\label{tab:readability_results}
\end{table}

\begin{table}[t]
\centering
\small
\begin{tabular}{llllll}
             & \textbf{R1}$\uparrow$ & \textbf{R2}$\uparrow$ & \textbf{RL}$\uparrow$ & \textbf{BLEU}$\uparrow$ & \textbf{SARI}$\uparrow$ \\
\midrule
No UL       & 0.40              & 0.15              & 0.37              & 0.44          & 0.38          \\
UL-Cochrane & 0.38              & 0.14              & 0.36              & 0.39          & 0.40          \\
UL-Newsela  & 0.39              & 0.15              & 0.37              & 0.43          & 0.39          \\
UL-Both     & 0.38              & 0.14              & 0.37              & 0.40          & 0.39          \\
\end{tabular}
\caption{ROUGE, BLEU, and SARI scores for generated PLS. All differences between No-UL and UL models, except for (BLEU, UL-Newsela), are statistically significant ($p<0.01$, paired $t$-test).}
\label{tab:rouge_bleu_results}
\end{table}

\subsection{Content}

We report SARI~\cite{xu2016optimizing}, a standard edit-based metric for text simplification, and BLEU~\cite{papineni2002bleu}, a precision-based method for machine translation that is also often reported for simplification systems.  \citet{xu2016optimizing} showed that SARI correlates better with human evaluation for simplification tasks, focusing more on simplicity, while BLEU is stronger with respect to meaning and grammar. Finally we report the F1 versions of ROUGE-1, ROUGE-2, and ROUGE-L~\cite{lin2004rouge}, which are the standard metrics typically used for summarization tasks.

Table \ref{tab:rouge_bleu_results} shows the mean ROUGE, BLEU, and SARI scores.
While UL models yielded small but significantly better SARI scores, the opposite is true for the ROUGE and BLEU measures. Despite the lack of clear patterns in these scores, there are clear qualitative differences between the different models' outputs, which are expounded upon in Section \ref{sec:qualitative_observations}.

\paragraph{Extractive vs.\ abstractive?} Although not reflected in the automatic evaluation metrics above, the increase in readability of UL models led us to suspect that UL models are more abstractive than extractive, namely, they contain more paraphrases. To determine the degree to which the outputs directly copy content from the technical abstracts, we computed the fraction of $n$-grams in the output PLS that also occur in the abstract (without considering repetition). These results are shown in Table \ref{tab:ngram_overlap_results}.

We observe that the introduction of UL clearly decreases $n$-gram overlap, and the difference becomes more marked as $n$ increases. 
The use of Cochrane weights (those from the logistic regression model trained to discriminate between technical abstracts and PLS) likely reduces $n$-gram overlap because the tokens most penalized in UL training are those used to represent numerical data, e.g., statistics and confidence intervals. 
Penalizing these tokens discourages the regurgitation of numerical details from the technical abstract. The use of Newsela weights does not have the same effect, again likely due to the domain difference between the tokens penalized during unlikelihood training and those generated by the model. None of the model settings, however, achieve $n$-gram overlap scores nearly as low as the reference PLS, indicating that the generated summaries remain considerably more extractive than human-written PLS. 

\begin{table}[t]
\centering
\small
\begin{tabular}{lllll}
& \textbf{N=1} & \textbf{N=2} & \textbf{N=3} & \textbf{N=4} \\ \midrule
\emph{PLS}   & 0.56         & 0.29         & 0.19         & 0.14         \\ \midrule
No-UL       & 0.95         & 0.89         & 0.84         & 0.79         \\
UL-Cochrane & 0.84         & 0.67         & 0.57         & 0.49         \\
UL-Newsela  & 0.92         & 0.81         & 0.73         & 0.66         \\
UL-Both     & 0.89         & 0.76         & 0.67         & 0.59         \\
\end{tabular}
\caption{\% of $n$-grams in reference/generated PLS that are also in the abstracts.}
\label{tab:ngram_overlap_results}
\end{table}

\subsection{Manual examination and analysis}~\label{sec:qualitative_observations}
We manually examined the outputs generated by our models on a random sample of 40 technical abstracts from the test split of our dataset. 
While reading these outputs, we made special note of text length, readability and coherence, the presence of hallucinated information not found in the corresponding abstract, and artifacts such as repetition and misspelled words.

Our examination demonstrated that the generated texts were all significantly shorter than their respective abstracts and also shorter than the reference PLS. 
Furthermore, the models trained with Cochrane weights (`UL-Cochrane' and `UL-Both') produced shorter texts on average than the models trained without UL or with Newsela weights. 
This observation is supported by the results in Table \ref{tab:length_results}, which displays the average number of tokens and sentences in the summaries generated under different training settings.

One explanation for why UL with Cochrane weights produces shorter summaries is that training with these weights discourages the copying of statistics from the original abstract, a phenomenon exemplified in Appendix A, Table \ref{tab:results_examples}. Another trend that we noticed was that higher $\alpha$ values produce shorter, more readable summaries at the expense of information completeness. Training with a high $\alpha$ also increases the likelihood of hallucination, misspelling, and repetition. These drawbacks greatly impacted coherence for $\alpha\geq 1000$.
These observations suggest a tradeoff between completness of information and conciseness as $\alpha$ is varied in the training process.

The most common hallucination found in all settings, and especially with high $\alpha$, was the inclusion of a statement of the form \emph{The evidence is current to [month] [year]}. The reason for this is that many PLS contain such a statement of currency not found in the technical abstracts, so models learn to include such a statement even if it cannot be factually deduced from the abstract. 
Another observation is that most commonly misspelled words are those of medications and diseases. 
Table ~\ref{tab:artifacts_examples} provides examples of the various kinds of artifacts found in the generated PLS. The presence of these artifacts suggest that in practice, generated texts should be reviewed before being used.

\begin{table}
\centering
\small
\begin{tabular}{p{7.5cm}}
\toprule
\textbf{Hallucination:}
\hl{The evidence is up-to-date as of February 2016.} We found seven studies, involving 1839 participants, that compared home-based treatment with hospital-based care for venous thromboembolism. \\
\textbf{Misspelling:} 
The review authors provided no information on other important outcomes, including gastro-oesophageal reflux, aspiration pneumonia, \hl{necrotise enterulitis}... \\
\textbf{Repetition:}
However, we were not able to combine their results because of the \hl{small number and small number} of people in the included studies.\\
\bottomrule
\end{tabular}
\caption{Example of artifacts found in generated PLS.}
\label{tab:artifacts_examples}
\end{table}

\begin{table}[t]
\centering
\small
\begin{tabular}{lllll}
                 & \textbf{\# Tokens} & \textbf{\# Sentences} \\
\midrule
\emph{Abstracts} & 492.04       & 14.03      \\
\emph{PLS}       & 254.60       & 9.59      \\
\midrule
No UL           & 228.27       & 8.34      \\
UL-Cochrane     & 163.79       & 7.10      \\
UL-Newsela      & 201.01       & 8.45      \\
UL-Both         & 173.88       & 7.75
\end{tabular}
\caption{Lengths of generated PLS.}
\label{tab:length_results}
\end{table}

\section{Conclusions}

In this work we considered the important task of medical text simplification. We derived a new resource for this task made up of technical abstracts summarizing medical evidence paired with plain language versions of the same; we have made this data publicly available to facilitate further research.\footnote{We emphasize that the data here comprises only text derived from publicly accessible abstracts.} We proposed a new masked language model (MLM)-based measure of the technicality of text, which quantifies technicality by calculating the likelihood of tokens in the input text with respect to a transformer-based MLM trained on a technical corpus. We demonstrated that this metric better discriminated technical abstracts from PLS than more traditional notions of readability.

We proposed models for automated simplification based on BART \cite{lewis2019bart}, extending the training objective by incorporating an explicit penalty for production of `jargon' terms. We found that this method can improve model outputs (i.e., can increase simplicity and the abstractiveness of summaries) according to the metrics considered.

\section{Ethical Considerations}
This paper presents a dataset from the Cochrane library; this comprises only the freely available portion of the information on Cochrane (abstracts that are readily available to all). No annotators other than the authors of this paper are involved in the manual inspection of this data. In addition, the Cochrane data in itself, and our collection and inspection of it, does not involve any personally identifiable information.

The baseline models presented involves simplifying medical texts. Inconsistencies (e.g., hallucinations) of the generated PLS with respect to the original review is an artifact discussed in Section~\ref{sec:qualitative_observations}. This can lead to misinformed readers. Therefore, the outputs of the proposed systems should always be manually examined before being used.

\section*{Acknowledgments}

This work was supported in part by the National Institutes of Health (NIH), grant R01-LM012086, and the National Science Foundation (NSF), grant IIS-1850153. We acknowledge the Texas Advanced Computing Center (TACC) at The University of Texas at Austin for providing HPC resources that have contributed to the research results reported within this paper.

\bibliography{acl2020}
\bibliographystyle{acl_natbib}

\clearpage
\appendix

\begin{table*}[t]
\centering
\small
\begin{tabular}{p{15cm}}
\textbf{\large Appendix A: Example outputs}\\
~\\
\toprule
\textbf{Technical abstract:} 
We included a total of 40 studies in the review, with more than 140,000 women aged between 20 and 70 years old. Many studies were at low risk of bias. There were a sufficient number of included studies with adequate methodology to perform the following test comparisons: hybrid capture 2 (HC2) (1 pg/mL threshold) versus conventional cytology (CC) (atypical squamous cells of undetermined significance (ASCUS)+ and low-grade squamous intraepithelial lesions (LSIL)+ thresholds) or liquid-based cytology (LBC) (ASCUS+ and LSIL+ thresholds), other high-risk HPV tests versus conventional cytology (ASCUS+ and LSIL+ thresholds) or LBC (ASCUS+ and LSIL+ thresholds). For CIN 2+, pooled sensitivity estimates for HC2, CC and LBC (ASCUS+) were 89.9\%, 62.5\% and 72.9\%, respectively, and pooled specificity estimates were 89.9\%, 96.6\%, and 90.3\%, respectively. The results did not differ by age of women (less than or greater than 30 years old), or in studies with verification bias. Accuracy of HC2 was, however, greater in European countries compared to other countries. The results for the sensitivity of the tests were heterogeneous ranging from 52\% to 94\% for LBC, and 61\% to 100\% for HC2. Overall, the quality of the evidence for the sensitivity of the tests was moderate, and high for the specificity. The relative sensitivity of HC2 versus CC for CIN 2+ was 1.52 (95\% CI: 1.24 to 1.86) and the relative specificity 0.94 (95\% CI: 0.92 to 0.96), and versus LBC for CIN 2+ was 1.18 (95\% CI: 1.10 to 1.26) and the relative specificity 0.96 (95\% CI: 0.95 to 0.97). The relative sensitivity of HC2 versus CC for CIN 3+ was 1.46 (95\% CI: 1.12 to 1.91) and the relative specificity 0.95 (95\% CI: 0.93 to 0.97). The relative sensitivity of HC2 versus LBC for CIN 3+ was 1.17 (95\% CI: 1.07 to 1.28) and the relative specificity 0.96 (95\% CI: 0.95 to 0.97). Whilst HPV tests are less likely to miss cases of CIN 2+ and CIN 3+, these tests do lead to more unnecessary referrals. However, a negative HPV test is more reassuring than a negative cytological test, as the cytological test has a greater chance of being falsely negative, which could lead to delays in receiving the appropriate treatment. Evidence from prospective longitudinal studies is needed to establish the relative clinical implications of these tests.\\ \midrule
\textbf{Plain-language summary:}
We searched for all relevant studies up to November 2015. Forty studies compared the HPV test to the Pap test on over 140,000 women between 20 to 70 years old who attended for their routine cervical screening. The studies examined which test can detect precancerous cervical changes which are called cervical intraepithelial neoplasias (CIN 2 and CIN 3). There were enough studies with enough women in them to allow us to draw conclusions. However, some of the results from the studies were different from each other. For example, tests were more accurate in studies in Europe than in Asia or Central or South America. Overall, the quality of the evidence was moderate to high. A perfect test would correctly say if a woman has precancerous changes or if a woman does not. But most tests are not perfect. This review found that for every 1000 women screened, around 20 women will have precancerous changes. The HPV test will correctly identify 18 of these women (but will miss 2 women). The Pap test will identify 15 of the women (but will miss 5 women). The women who are missed could develop cervical cancer. For every 1000 women screened, there will be 980 women who will not have precancerous changes. The HPV test will correctly identify 881 women (but 99 women will be incorrectly told that they have a lesion). The Pap test will correctly identify 885 women (but 95 will be incorrectly told that they have a lesion). Women who are incorrectly told that they have a lesion may have their cervix examined or may receive surgery unnecessarily.\\ \midrule
\textbf{No UL}: 
We included 40 studies in the review, with more than 140,000 women aged between 20 and 70 years old. Many studies were at low risk of bias. There were a sufficient number of included studies with adequate methodology to perform the following test comparisons: hybrid capture 2 (HC2) (1 pg/mL threshold) versus conventional cytology (atypical squamous cells of undetermined significance (ASCUS)+ and low-grade squamous intraepithelial lesions (LSIL)+ thresholds) or liquid-based cytology or other high-risk HPV tests versus conventional or liquid cytology thresholds. For CIN 2+, pooled sensitivity estimates for HC2, CC and LBC were 89.9\%, 96.6\%, and 90.3\%, respectively. The results did not differ by age of women (less than or greater than 30 years old), or in studies with verification bias. The accuracy of HC2 was, however, greater in European countries compared to other countries. Overall, the quality of the evidence for the sensitivity of the tests was moderate, and high for the specificity. Whilst HPV tests are less likely to miss cases of CIN2+ and CIN 3+, these tests do lead to more unnecessary referrals. However, a negative HPV test is more reassuring than a negative cytological test, as the cytology test has a greater chance of being falsely negative, which could lead to delays in receiving the appropriate treatment. Evidence from prospective longitudinal studies is needed to establish the relative clinical implications of these tests.\\ \midrule
\textbf{UL, Cochrane Weights, Temp=2, $\alpha$=100}:
We found 40 studies involving more than 140,000 women aged between 20 and 70 years old. The studies compared different types of HPV tests with each other, or with conventional cytology. The results showed that the tests were more accurate than conventional cytological tests. However, the results did not differ by age of women (less than or greater than 30 years old), or in studies with verification bias. Overall, the quality of the evidence for the sensitivity of the tests was moderate, and the specificity was high for the tests. The accuracy of the test was, however, greater in European countries compared to other countries. Whilst HPV tests are less likely to miss cases, they do lead to more unnecessary referrals, as the cytological test has a greater chance of being falsely negative, which could lead to delays in receiving the appropriate treatment. Evidence from prospective longitudinal studies is needed to establish the relative clinical implications of these tests.\\
\bottomrule
\end{tabular}
\caption{A full example of technical abstract, reference PLS and model outputs.}
\label{tab:results_examples}
\end{table*}

\end{document}